\documentclass[10pt,twocolumn,letterpaper]{article}

 \usepackage[pagenumbers]{cvpr} 

\usepackage[accsupp]{axessibility}
\usepackage[utf8]{inputenc} 
\usepackage[T1]{fontenc}    
\usepackage{url}            
\usepackage{booktabs}       
\usepackage{amsfonts}       
\usepackage{nicefrac}       
\usepackage{microtype}      
\usepackage{bbm}

\usepackage{graphicx}
\usepackage{amsmath}
\usepackage{amssymb}
\usepackage{multirow}
\usepackage{booktabs}
\usepackage{tabularx}
\usepackage[dvipsnames,table,xcdraw]{xcolor}
\usepackage{overpic}
\usepackage{colortbl}
\usepackage{makecell}
\usepackage{contour}

\newcommand{\figref}[1]{Fig.~\ref{#1}}
\newcommand{\tabref}[1]{Tab.~\ref{#1}}

\newcommand{\secref}[1]{Sec.~\ref{#1}}

\definecolor{cvprblue}{rgb}{0.21,0.49,0.74}
\usepackage[pagebackref,breaklinks,colorlinks,linkcolor=BrickRed,citecolor=cvprblue]{hyperref}

\usepackage{enumitem}
\usepackage{pifont}

\newcommand{\myParaP}[1]{\vspace{.05in}\noindent\textbf{#1}}
\newcommand{\myPara}[1]{\vspace{.05in}\noindent\textbf{#1:}}

\newcommand{\highlight}[1]{\textbf{\textcolor{ForestGreen}{#1}}}

\def\paperID{1601} 
\def\confName{CVPR}
\def\confYear{2024}

\graphicspath{{../figure/}, {../figure/Related/}}

\begin{document}

\title{Traffic Scene Parsing through the TSP6K Dataset}

\author{
  Peng-Tao Jiang$^{1,2}$\thanks{The first two authors contributed equally to this work. Part of this work was done when P.-T. Jiang was a postdoc researcher at Zhejiang University.} ~~ Yuqi Yang$^{1*}$ ~~ Yang Cao$^3$ ~~ Qibin Hou$^{1,4}$\thanks{Q. Hou is the corresponding author. 
  }  ~~ Ming-Ming Cheng$^{1,4}$ ~~ Chunhua Shen$^2$ \\
  [0.2cm]
$^1$VCIP, CS, Nankai University \quad
$^2$Zhejiang University \quad 
$^3$HKUST \quad 
$^4$NKIARI, Shenzhen Futian\\
{\tt\small $\tt pt.jiang@mail.nankai.edu.cn$,
$\tt yangyq2000@mail.nankai.edu.cn$,
$\tt andrewhoux@gmail.com$ }}

\maketitle

\begin{abstract}
Traffic scene perception in computer vision is a critically important task 
to achieve intelligent cities.  
To date, most existing datasets focus on autonomous driving scenes. 
We observe that the models trained on those driving datasets 
often yield unsatisfactory results on traffic monitoring scenes.
However, little effort has been put into improving the traffic monitoring scene 
understanding, mainly due to the lack of specific datasets.  
To fill this gap, we introduce a specialized traffic monitoring  dataset, 
termed TSP6K, containing images from the traffic monitoring scenario, 
with high-quality pixel-level and instance-level annotations.
The TSP6K dataset captures more crowded traffic scenes 
with several times more traffic participants than the existing driving scenes. 
We perform a detailed analysis of the dataset and comprehensively 
evaluate previous popular scene parsing methods, instance segmentation methods 
and unsupervised domain adaption methods.
Furthermore, considering the vast difference in instance sizes, 
we propose a detail refining decoder for scene parsing, 
which recovers the details of different semantic regions 
in traffic scenes owing to the proposed TSP6K dataset.
Experiments show its effectiveness in parsing the traffic monitoring scenes.
Code and dataset are available at \url{https://github.com/PengtaoJiang/TSP6K}.
\end{abstract}

\section{Introduction} \label{sec:intro}
As a classic and important computer vision task, the scene parsing task 
aims to segment the semantic objects and stuff from the given images. 
Nowadays, with the emergence of large-scale scene understanding datasets, 
such as ADE20K~\cite{zhou2017scene} and COCO-Stuff \cite{caesar2018cocostuff}, 
has greatly promoted the development of scene understanding algorithms \cite{long2015fully,zhao2016pyramid,lin2016refinenet,xie2021segformer,zheng2021rethinking,jiang2023deep,du2022MEGFNet}.  
Many application scenarios, such as robot navigation \cite{humblot2022navigation,crespo2020semantic}
and medical diagnosis \cite{ronneberger2015u}, 
benefit from the advanced scene understanding algorithms.
As an important case of scene understanding, traffic scene understanding 
focuses on understanding urban street scenes, where the most frequently 
appeared instances are humans, vehicles, and traffic signs. 
To date, there are already many large-scale publicly available traffic scene datasets, 
such as KITTI~\cite{geiger2013vision}, Cityscapes~\cite{cordts2016cityscapes}, 
and BDD100K \cite{yu2020bdd100k}.
Benefiting from these finely anontated datasets, the segmentation 
performance of the recent scene understanding approaches~\cite{zhang2017scale,zhu2019asymmetric,he2019adaptive,liang2018dynamic,strudel2021segmenter,cheng2019spgnet,wu2022yolop,li2022ar} 
has also been considerably improved.
%


\newcommand{\addTexB}[1]{\contour{white}{\textcolor{black}{#1}}}
\newcommand{\addFigT}[2]{\begin{overpic}[width=0.161\textwidth,height=0.10\textwidth]{figs/example_label/#1}\put(0.3,4){\addTexB{#2}}\end{overpic}}
\newcommand{\addFigsT}[2]{\addFigT{#1.jpg}{#2} & \addFigT{#1_sem.png}{} & \addFigT{#1_ins.png}{}  }
\begin{figure*}[t]
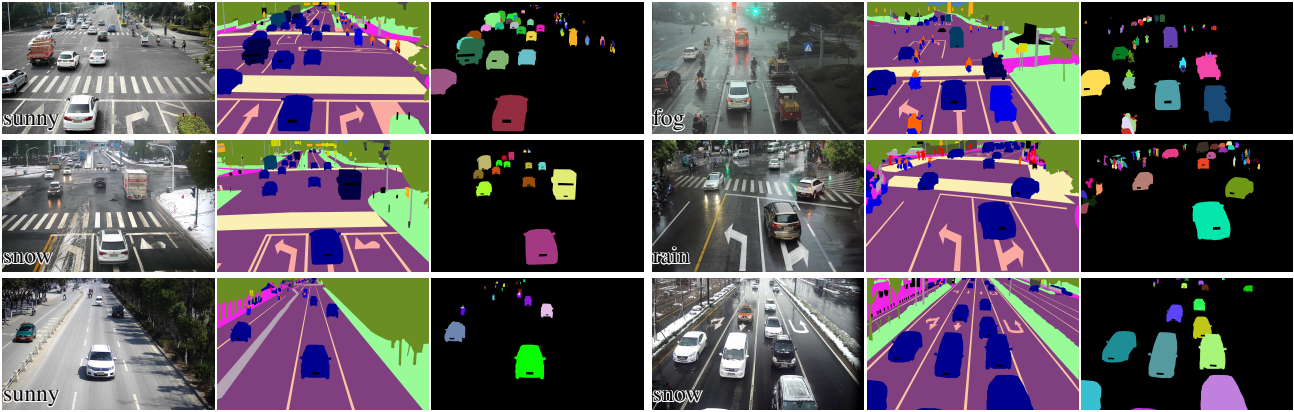
 
\centering
\small
\setlength\tabcolsep{0.6pt}
\renewcommand{\arraystretch}{0.8}
\begin{tabular}{cccccc}
  \addFigsT{traffic_00023}{sunny} & \addFigsT{traffic_42204}{fog} \\
  \addFigsT{traffic_41397}{snow}  & \addFigsT{traffic_44857}{rain} \\
  \addFigsT{traffic_46231}{sunny}  & \addFigsT{traffic_43991}{snow} \\
\end{tabular}
\vspace{-5pt}
\caption{Examples are randomly picked from the TSP6K dataset.
Each image is associated with its corresponding semantic label and instance label. 
We have masked the vehicle plates for privacy protection. 
}\label{fig:examp}
\vspace{-5pt}
\end{figure*}

%
A characteristic of these traffic datasets is that they are 
mostly collected from a driving platform, 
such as a driving car, and hence are more suitable 
for the autonomous driving scenario. 
However, little attention has been 
spent on the traffic monitoring scenes.
%
Traffic 
monitoring  scenes are usually captured by the shooting platform 
high-hanging (4.5-6 meters) 
in 
the street, 
which offers a rich vein of information on traffic flow~\cite{lv2014traffic,jin2017predicting}.
The high-hanging shooting platform usually 
observes more traffic participants 
than the driving ones, especially at the crossing.
We observe that deep learning models trained on these existing 
traffic datasets 
can obtain poor results in parsing traffic monitoring 
scenes, possibly because of the domain gap.
Although analyzing the traffic monitoring scenes is in demand for 
many applications, such as traffic flow analysis \cite{lv2014traffic,jin2017predicting}, 
no current traffic scene datasets are available for facilitating such research, 
to the best of our knowledge.

To facilitate the research on parsing the traffic monitoring scenes, 
we construct a specific dataset for traffic scene analysis and present it 
in this paper.
Specifically, we carefully collect many traffic images 
from the urban road shooting platforms at different locations.
To keep the diversity of our dataset, we collect images from hundreds of 
traffic scenes at different times of the day.
To conduct semantic segmentation and instance segmentation on this dataset, 
we ask annotators to finely annotate them with high-quality semantic 
and instance-level labels.
Due to the expensive labor for annotations, we finally obtained 6,000 
finely anontated traffic images.
In \figref{fig:examp}, we have shown some traffic images 
and their corresponding semantic-level and instance-level labels.
Using these finely anontated labels, we perform a comprehensive study 
about the traffic monitoring scenes.
The characteristics of this dataset are summarized as follows: 
\textbf{1)} the largest traffic monitoring datasets, \textbf{2)} much more crowded scenes, 
\textbf{3)} wide variance of instance sizes, 
and \textbf{4)} a large domain gap between the driving scenes and the monitoring scenes.
Based on the proposed TSP6K dataset, we evaluate a few classic scene parsing 
methods, instance segmentation methods, and unsupervised domain adaption methods.
We show and analyze the performance of different methods on the proposed TSP6K dataset.

In addition, we propose a detail refining decoder for image segmentation on TSP6K. 
The detail refining decoder utilizes the encoder-decoder structure 
and refines the high-resolution features with a region refining module.
The region refining module utilizes the attention mechanism 
and computes the attention between the pixels and each region token.
The attention is further used to refine the pixel relationships 
in different semantic regions.
Taking the backbone of the SegNeXt \cite{guo2022segnext} as our encoder, 
our proposed detail refining decoder method achieves 75.8\% mIoU score and 58.4\% iIoU score on the TSP6K validation set, which are 1.2\% and 1.1\% higher 
than those of the state-of-the-art SegNeXt work.
To summarize, our main contributions are as follows:
\begin{itemize}
    \item We propose a specialized traffic dataset for researching 
    the task of traffic monitoring scene parsing, termed TSP6K, 
    which collects images spanning various scenes 
    from the urban road shooting platform. 
    We also provide pixel-level annotations of semantic labels 
    and instance labels. 
    \vspace{1pt}
    
    \item Based on the TSP6K dataset, we evaluate the performance of 
    previous scene parsing methods and instance segmentation methods 
    on traffic monitoring scenes.  
    Moreover, the TSP6K dataset can also serve as an additional supplement 
    for evaluating unsupervised domain adaption methods 
    for traffic monitoring applications. 
    \vspace{1pt}

    \item To improve traffic monitoring scene parsing, 
    we propose a detail refining decoder that learns region 
    tokens to refine different regions on high-resolution features.
    Experiments validate the effectiveness of the proposed decoder.
\end{itemize}

\section{Related Work} \label{sec:related}

\subsection{Scene Parsing Datasets}
Scene parsing datasets with full pixel-wise annotations are utilized for training 
and evaluating the scene parsing algorithms.
As an early one, the PASCAL VOC dataset~\cite{everingham2015pascal} 
was proposed in a challenge that aims to parse the objects of 20 carefully 
selected classes in each image.
Later, the community proposed more complex datasets with many more classes, 
such as COCO~\cite{lin2014microsoft} and ADE20K~\cite{zhou2017scene}.
The scenes in the above datasets span a wide range.
Different from these datasets, there are also some datasets focusing on 
particular scenes, such as the traffic scenes.
There exist many traffic scene parsing datasets \cite{huang2019apolloscape,zendel2018wilddash,neuhold2017mapillary,zendel2022unifying,sakaridis2019guided,dai2018dark,varma2019idd}, 
such as KITTI~\cite{geiger2013vision}, Cityscapes~\cite{cordts2016cityscapes}, 
ACDC \cite{sakaridis2021acdc}, and BDD100K~\cite{yu2020bdd100k}.
These traffic-parsing datasets annotate the most frequent 
classes in the traffic scenes, such as the traffic sign, 
rider, and vehicles, \etc
Based on these finely annotated traffic datasets, 
the approaches based on the neural networks have achieved 
great success in parsing traffic scenes.

Despite the success of the above datasets, we find that the traffic 
scenes in these datasets all from the driving platform.  
Models trained on these datasets often behave not well on 
the traffic monitoring scenes, which play an important role 
in traffic flow analysis.
In addition, the monitoring scenes usually capture much more traffic 
participants than the driving scenes.
The proposed TSP6K dataset is different from the driving datasets, 
aiming at improving the performance of scene parsing models on monitoring scenes, 
which can be regarded as a supplement to the current traffic datasets.
Furthermore, Kirillov \etal \cite{kirillov2023segment} has proposed 
a large segmentation dataset, SA-1B, containing 1 billion masks.
SA-1B also contains some monitoring traffic images.
However, the segmentation masks in SA-1B are all class-agnostic.
In contrast to that, the segmentation masks in our dataset 
are all class-known.

\subsection{Scene Parsing Approaches}
Convolutional neural networks have greatly facilitated the development 
of scene parsing approaches.
Typically, Long \etal \cite{long2015fully} first proposed
a fully convolutional network (FCN) that generates dense 
predictions for scene parsing.
Later, some approaches, such as the popular 
DeepLab \cite{chen2014semantic,chen2017deeplab} and PSPNet \cite{zhao2016pyramid},  
benefit from large receptive fields and multi-scale features, 
improving the performance by a large margin.
Besides, there are also some approaches~\cite{badrinarayanan2017segnet,lin2016refinenet,chen2018encoder,cheng2019spgnet}, 
utilizing the encoder-decoder structure to refine 
the low-resolution coarse predictions with the details 
of high-resolution features.
Except for the simple convolutions, researchers \cite{zhao2018psanet,yuan2021ocnet,hou2020strip,huang2019ccnet} 
found that the attention mechanism~\cite{guo2022attention} can significantly 
improve scene parsing networks due to their ability to 
model long-range dependencies.
In addition, there are also some works \cite{zhao2018icnet,yu2018bisenet,yu2021bisenet,zhang2022topformer,wu2019fastfcn} 
exploring real-time scene parsing algorithms, which take advantage of 
self-attention in efficient ways.

Recently, with the successful introduction of Transformers into image 
recognition~\cite{dosovitskiy2020image}, researchers 
have attempted to apply Transformers to the segmentation 
task~\cite{strudel2021segmenter,xie2021segformer,zheng2021rethinking,cheng2021per,cheng2022masked}.
Interestingly, some recent works~\cite{guo2022segnext,hou2022conv2former,guo2022visual} 
show that convolutional neural networks can perform better than Transformer-based models 
on the scene parsing task.
In our dataset, we also observe similar results.
SegNeXt~\cite{guo2022segnext} achieves the best performance on our TSP6K dataset 
using even fewer parameters than other works.
The proposed method in this work also adopts the backbone of the SegNeXt work.
But different from it, we design a detail refining decoder, which is more suitable 
for processing high-resolution images than the one used in SegNeXt.

\begin{figure} [t]
  \begin{subfigure}{.47\textwidth}
    \centering
    \includegraphics[width=1\linewidth]{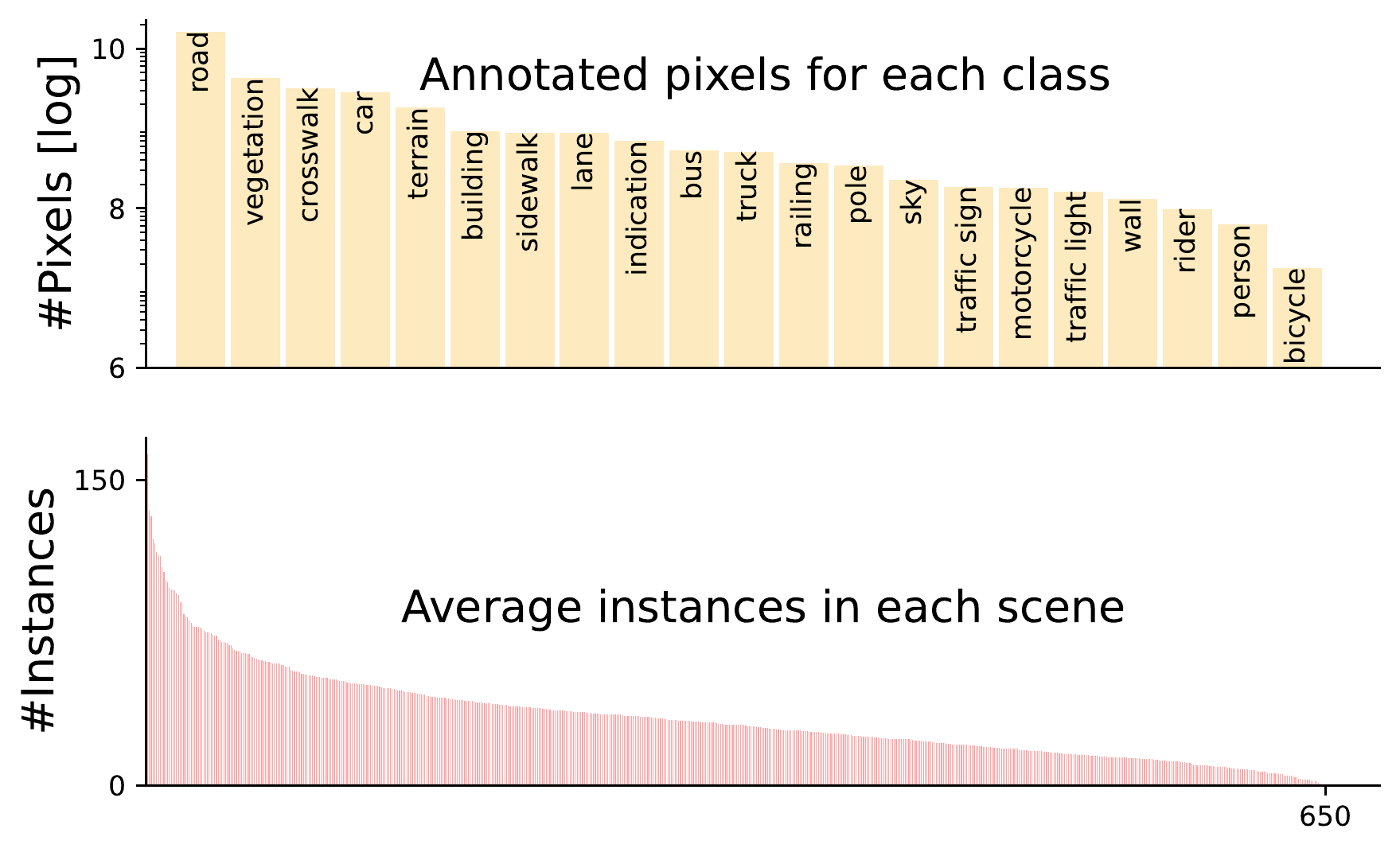}
    \subcaption{}
\end{subfigure}
  \begin{subfigure}{.47\textwidth}
    \centering
    \includegraphics[width=1\linewidth]{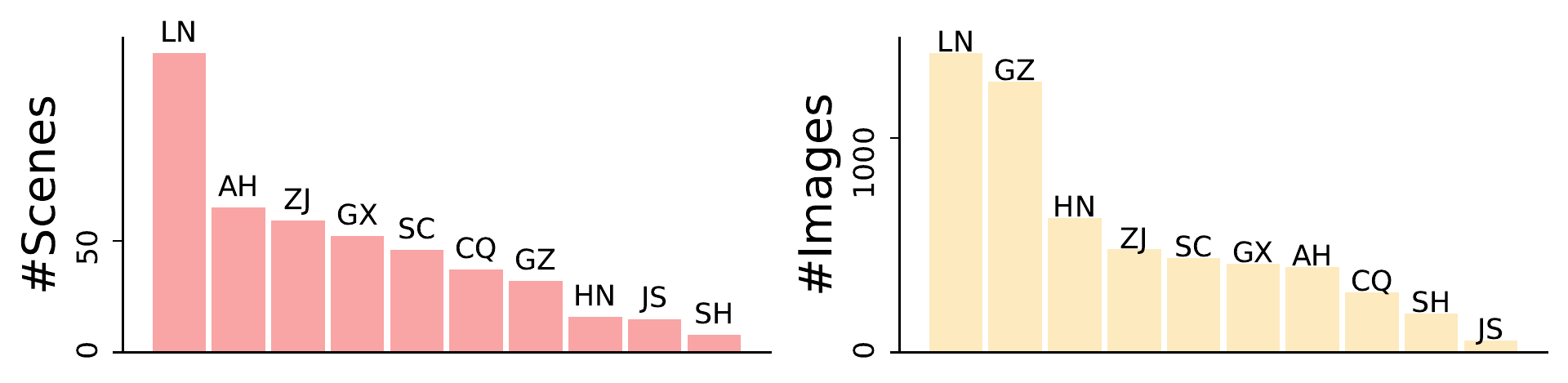}
    \subcaption{}
\end{subfigure}
  \vspace{-5pt}
  \caption{(a) Class and scene information of the TSP6K dataset. 
  (b) The geographic distribution of the scene and image. 
  }\label{fig:inst_per_scene}
  \vspace{-5pt}
\end{figure}

\subsection{Instance Segmentation Approaches}
Instance segmentation aims to segment and distinguish 
each instance of the same class.
Previous instance segmentation approaches can be roughly 
divided into two groups, box-dependent methods \cite{he2017mask,fang2021instances,huang2019mask,chen2019hybrid,bolya2019yolact,liu2018path}, 
and box-free methods \cite{wang2020solo,wang2020solov2,xie2020polarmask,tian2020conditional}.
Box-dependent methods first detect the bounding box of 
the target object, and then perform binary segmentation in the 
box region. 
In contrast, box-free methods directly generate the 
instance mask for each instance and classify it in parallel.
In this paper, we select several methods for each 
category and evaluate them on TSP6K.

\subsection{Unsupervised Domain Adaption Approaches}
Unsupervised domain adaption (UDA) aims to adapt the models 
trained on one domain (with segmentation labels) to a new domain (without 
segmentation labels).
In recent years, lots of UDA approaches~\cite{tsai2018learning,hoffman2018cycada,vu2019advent,yang2020fda} 
for scene parsing emerge to address the domain discrepancy.
The UDA approaches mainly fall into two categories: 
adversarial training-based methods~\cite{hong2018conditional,vu2019advent,hoffman2018cycada,tsai2018learning,tsai2019domain,wang2020classes}, 
and self-training-based methods~\cite{lian2019constructing,zhang2019category,melas2021pixmatch,zou2018unsupervised,li2022class,zou2019confidence,wang2021uncertainty}.
The adversarial training-based methods attempt to 
align the feature representations or network predictions 
of the source and target domains.
The self-training-based methods generate pseudo masks 
for the target domain to train segmentation networks.
Previous UDA methods are usually evaluated by adapting from  
the synthetic traffic datasets (GTA5~\cite{richter2016playing} or 
SYNTHIA~\cite{ros2016synthia}) to the real traffic datasets (Cityscapes~\cite{cordts2016cityscapes}).
In this paper, we evaluate the UDA methods for adapting 
both the synthetic and real driving scenes (SYNTHIA~\cite{ros2016synthia}), 
Cityscapes~\cite{cordts2016cityscapes}) to the monitoring scenes (TSP6K).

\begin{table*}[t]
    \small
    \centering
    \caption{Comparisons among different traffic scene parsing datasets. 
    \textbf{Avg TP} denotes the number of the average traffic participants 
    in each image.
    \textbf{TP $>$ 50} denotes the number of images that contain 
    more than 50 traffic participants.
    As the instance labels of the test sets in other datasets 
    are not available, we all count the traffic participants 
    in the training and validation sets.  We can see that 
    our TSP6K dataset contains more traffic images with 
    more than 50 traffic participants when compared with other 
    datasets.
    }
    \label{tab:comp_dataset}
    \vspace{-5pt}
    \renewcommand{\arraystretch}{1.1}
    \setlength\tabcolsep{1.2mm}
    \begin{tabular}{crrrrrrrrrrr} \toprule[1.0pt]
       Type  & Datasets  &  Class & Weather  & \#Images  & Resolution & Annotation &  Avg TP   &  TP $>$ 50   & TP $>$ 75 & TP $>$ 100\\ 
       \midrule[0.8pt]
       \multirow{6}{*}{Driving} 
         & KITTI~\cite{geiger2013vision}              & 19  & Good    &  7,481  & 1,241$\times$375   &Pixel\&Inst & 4.9   & 0    & 0  &  0\\  
       ~ & Cityscapes~\cite{cordts2016cityscapes}     & 19  & Good    &  5,000  & 2,048$\times$1,024 & Pixel\&Inst & 18.8  & 54   & 10 &  4\\
       ~ & WildDash 2~\cite{zendel2018wilddash}       & 26  & Diverse & 5,068   & 1920$\times$1080   & Pixel\&Inst & 9.0   & 12   & 4  &  2 \\
       ~ & Mapillary~\cite{neuhold2017mapillary}      & 65  & Diverse & 25,000  & 3,436$\times$2,486 & Pixel\&Inst & 12.3  & 102  & 15 &  3\\
       ~ & ACDC~\cite{sakaridis2021acdc}              & 19  & Diverse & 3,142   & 1080$\times$1920   & Pixel\&Inst & 6.3   & 0    & 0  &  0\\
       ~ & BDD100K~\cite{yu2020bdd100k}               & 40  & Good    & 10,000  & 1,280$\times$720   & Pixel\&Inst & 12.8  & 5    & 0  &  0\\ \hline 
       \multirow{4}{*}{Monitoring} 
        & UrbanTracker~\cite{jodoin2014urban}         & 7   &  Good   &5(videos)& 1,035$\times$632   &  Box   &  3.7   &   0   &  0 &  0  \\  
       ~& CityFlow~\cite{tang2019cityflow}            & 1   &  Good   &40(videos)& 540$\times$960     &  Box   & 2.0    &   -   & -  &   -  \\   
       ~& AAU RainSnow~\cite{aau-rainbow}             & 3   &  Diverse &22(videos)& 640$\times$480     &  Pixel &  6.6   &   0   &  0 &   0  \\
\rowcolor{gray!16} ~ & TSP6K (ours)                   & 21  & Diverse & 6,000  & 2,942$\times$1,989 &  Pixel\&Inst & 42.0    & 1,227 & 367& 73\\
    \bottomrule[1.0pt]
    \end{tabular}  
    \vspace{-2pt}
\end{table*}

\begin{table}[ht]
    \small
    \centering
    \caption{Statistics of traffic participants in traffic images. 
    `\#H.' and `\#V.' denote \#Humans and \#Vehicles, respectively.}
    \label{tab:stp_dataset}
    \vspace{-5pt}
    \renewcommand{\arraystretch}{1.1}
    \setlength\tabcolsep{0.8mm}
    \begin{tabular}{l|c|c|c|c} \toprule[1.0pt]
       Datasets   &  \thead{\#Humans \\  $\left [ \mbox{10}^{3} \right ] $ } &  \thead{\#Vehicles \\  $\left [ \mbox{10}^{3} \right ] $ }  &  \#H./images  &  \#V./images \\ 
       \midrule[0.8pt]
       KITTI~\cite{geiger2013vision}           &   6.1    & 30.3  & 0.8  & 4.1  \\  
       Cityscapes~\cite{cordts2016cityscapes}  &   24.4   & 41.0  & 7.0  & 11.8  \\
       Mapillary~\cite{neuhold2017mapillary}   &   6.7    & 17.8  & 3.4  & 8.9   \\
       Wilddash2~\cite{zendel2018wilddash}     &   11.6   & 26.8  & 2.7  & 6.3   \\
       ACDC~\cite{sakaridis2021acdc}           &   3.8    & 15.9  & 1.2  & 5.1   \\  
       BDD100K~\cite{yu2020bdd100k}            &   11.7   & 90.3  & 1.5  & 11.3  \\
       \rowcolor{gray!16} TSP6K (ours)         &   64.0   & 188.2 & 10.7 & 31.3  \\
    \bottomrule[1.0pt]
    \end{tabular}  
    \vspace{-4pt}
\end{table}

\section{Dataset and Analysis} \label{sec:dataset}
In this section, we introduce the details for constructing the monitoring dataset 
and perform a comprehensive analysis of the proposed TSP6K dataset.
\subsection{Data Collection}
One significant aspect of researching the traffic monitoring scenes 
is data.
Once we construct a dataset for the monitoring scenes, 
the community researchers can improve the scene parsing 
results based on the novel data characteristics.
To facilitate the research, we aim to build a dataset 
specifically for researching the traffic monitoring scenes 
by collecting a large number of images from the high-hanging shooting platforms 
on different streets.
To ensure data diversity, the collection locations 
and weather conditions are highly considered.
Specifically, we collect the traffic images from about 10 
Chinese provinces with more than 600 scenes.
In \figref{fig:inst_per_scene}(b), we have shown the geographic 
distribution of scenes and images.
As the crossing and pedestrian crossing are an essential part of traffic scenes, 
where congestion and accidents often occur, we keep a majority of 
the traffic scenes containing the crossing. 
Besides, considering the weather diversity, we select the 
traffic images under various weather conditions, including 
sunny and cloudy day, rain, fog, and snow.
As a result, we finally selected 6,000 traffic images.

\subsection{Data Annotation}
After collecting data, we start to annotate the traffic images.
The complete annotated classes are shown in \figref{fig:inst_per_scene}(a).
Specifically, we annotate 21 classes, where most of the classes are 
the same as the class definition in Cityscapes~\cite{cordts2016cityscapes}.
We remove the unseen class `train' in our dataset 
and add three new classes.
As the indications on the road are vital for understanding the monitoring scenes, we ask the annotators to label three indication classes for traffic, namely crosswalks, driving indications, and lanes.
Besides, we have annotated the instance mask for each traffic participant.

Similar to the annotation policy of Cityscapes~\cite{cordts2016cityscapes}, 
the traffic images are also annotated from back to front.
To keep the quality of the labels, we design a double-check mechanism.
Specifically, the images are split into 30 groups, each of which contains 200 images.
When the annotators finish labeling the images, we pick 
30\% of 200 images and check if there exist class labeling errors.
If there exist class labeling errors in the selected images, 
we ask the annotator to check all the images in this group 
until there are no class labeling errors.

\subsection{Data Split}
The dataset is divided into three splits for training, 
validation, and test according to the ratio of 5:2:3. 
Images collected from different scenes are randomly 
split into different sets.
In total, there are 2,999, 1,207, and 1,794 images 
for the training, validation, and test sets, respectively.

\begin{figure*} [t]
  \begin{subfigure}{.35\textwidth}
    \centering
    \includegraphics[width=1\linewidth]{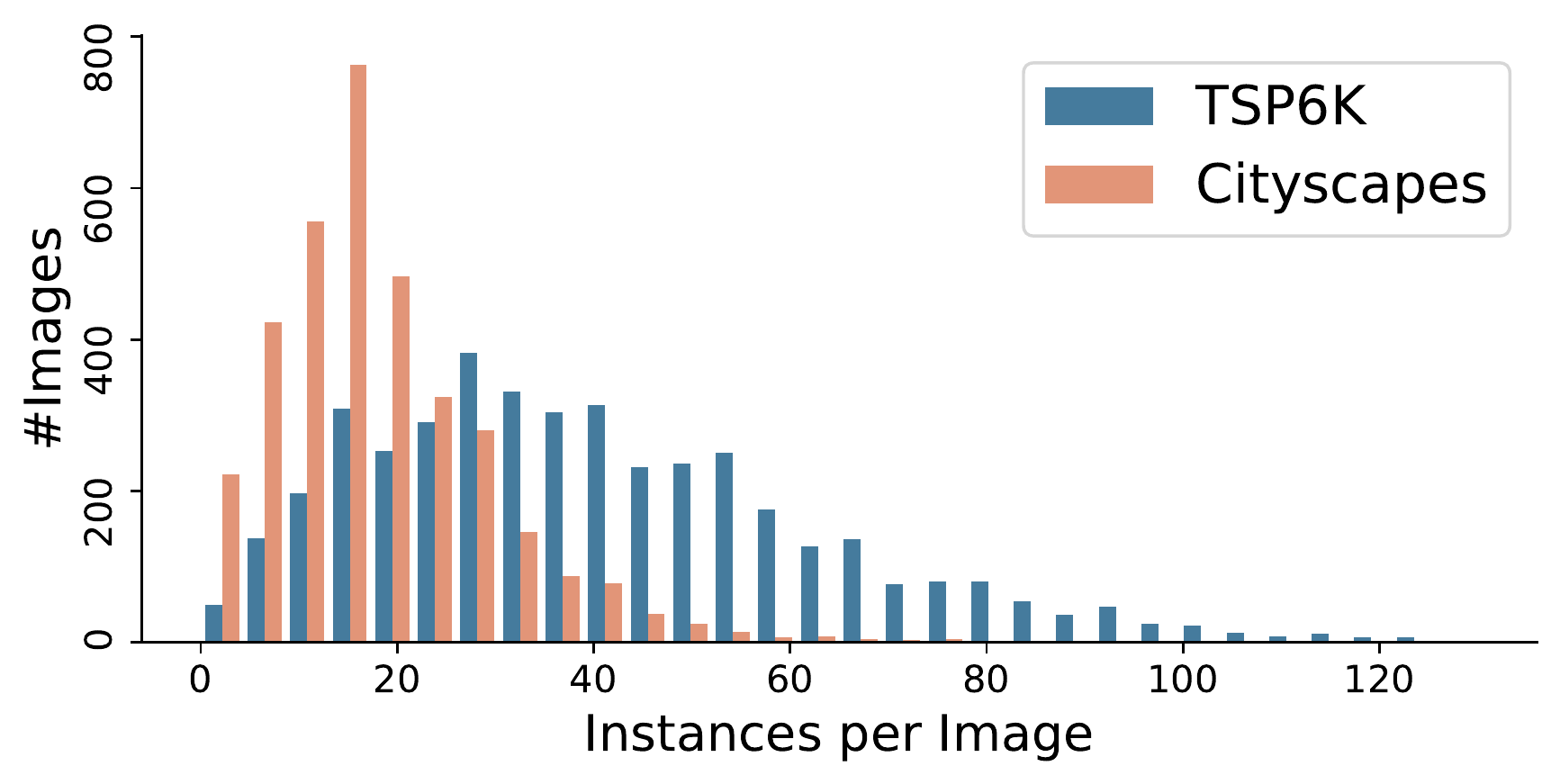}
    \subcaption{}
\end{subfigure}
  \begin{subfigure}{.35\textwidth}
    \centering
    \includegraphics[width=1\linewidth]{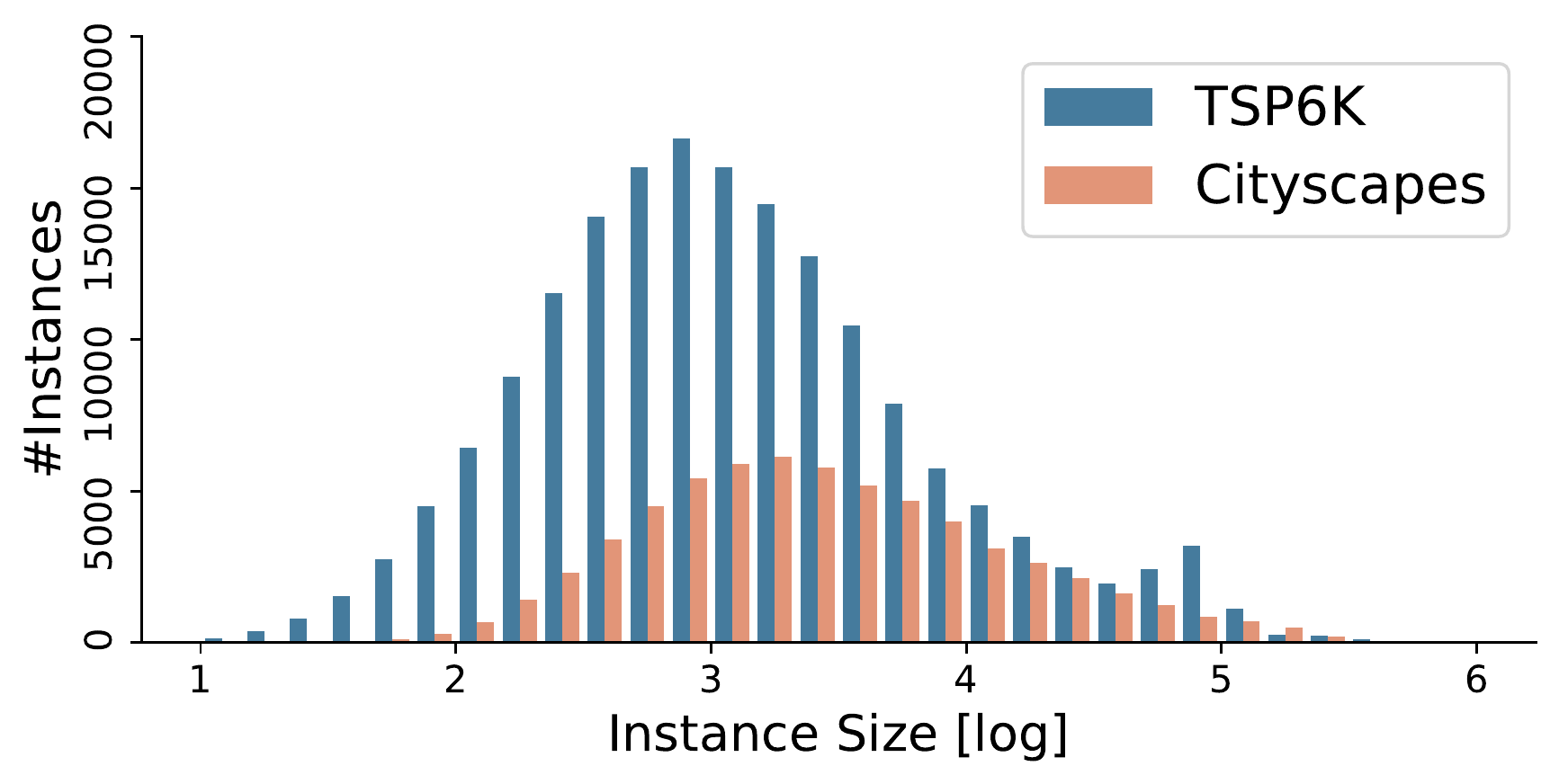}
    \subcaption{}
\end{subfigure}
  \begin{subfigure}{.28\textwidth}
    \centering
    \includegraphics[width=1\linewidth]{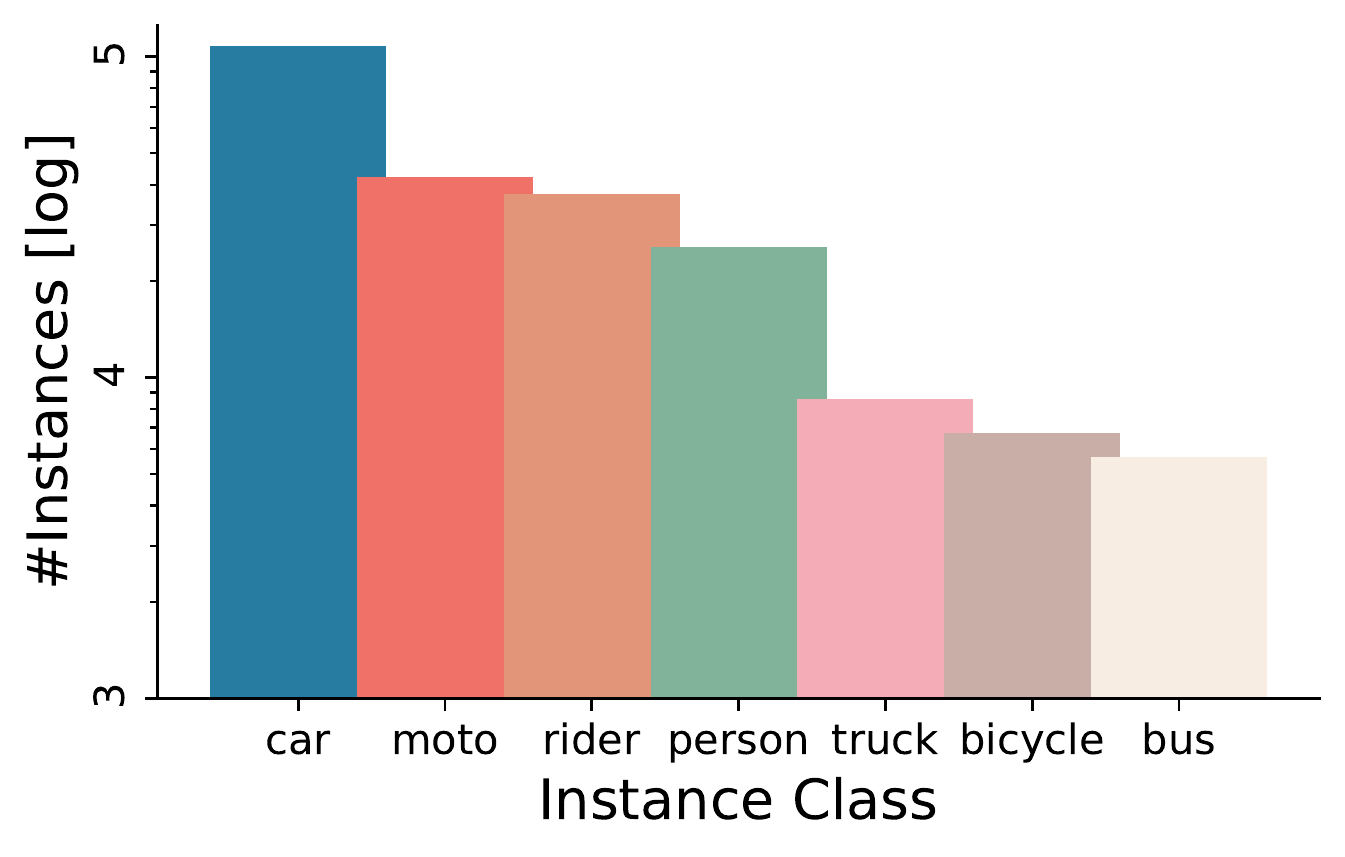}
    \subcaption{}
\end{subfigure}
  \vspace{-5pt}
\caption{
Data analysis of the TSP6K dataset. (a) The distribution of 
    the number of instances in each image. (b) The distribution of 
    the instance sizes.  (c) The number of instances 
    for each category. 
  }\label{fig:inst_per_img}
  \vspace{-5pt}
\end{figure*}

\subsection{Data Analysis}
%
%
We compare the TSP6K dataset with previous traffic datasets 
regarding the scene type, instance density, 
scale variance of instances, and domain gaps.
In \tabref{tab:comp_dataset} and \tabref{tab:stp_dataset}, 
we have listed the comparison among different traffic datasets.
The characteristics of TSP6K can be summarized 
as follows:

\myParaP{Largest traffic monitoring dataset:} 
To the best of our knowledge, most previous popular 
traffic datasets focus on the driving scenes.
The images of these datasets are collected from the driving platform.
There are also several datasets \cite{jodoin2014urban,tang2019cityflow} 
including the traffic monitoring scenes, shown in \tabref{tab:comp_dataset}.
However, they mainly focus on traffic participant tracking 
and only provide class-agnostic bounding box annotations.
Different from them, we address traffic monitoring scene parsing 
and provide semantic and instance annotations.
Compared with the existing traffic monitoring datasets, 
TSP6K contains much more labeled semantic classes, 
available instance segmentation, larger image resolution 
and a much larger number of images.

\myParaP{Much more crowded scenes:} 
One of the most important characteristics is that the TSP6K dataset 
contains more crowded images than those in driving datasets.
Since the majority of the traffic scenes are shooted at the crossing, 
the instance density on the road is much larger than the driving scenes.
In \tabref{tab:comp_dataset}, it can be seen that the driving 
datasets have few images containing more than 50 instances.
In contrast, our TSP6K dataset has a large number of images containing 
more than 50 instances, occupying about 30\% of the images 
in the training and validation sets.
Moreover, as shown in \tabref{tab:stp_dataset}, 
there are 10.7 humans and 31.3 vehicles on average in TSP6K, 
which exceeds other driving datasets several times. 
In addition, it can be seen that the existing monitoring datasets 
contain fewer annotated instances than driving datasets.
This can be mainly attributed to the incomplete annotation 
where only the moving vehicles are annotated.

\myPara{Wide variance in the instance sizes} 
For the monitoring scenes, the scale difference 
of the instances in the front and end is very large. 
As shown in \figref{fig:inst_per_img}(b), the instance size of TSP6K 
spans a wider range than Cityscapes.
Furthermore, TSP6K also contains more small traffic instances than Cityscapes.
The high-hanging platform usually has a much broader view 
than the driving platform. 
Thus, it can capture much more content in the distance. 
The huge variance of the instance sizes shows real traffic scenarios.

\myPara{Large domain gap} 
There exists a large domain gap between TSP6K and Cityscapes/BDD100K. 
The models trained on driving datasets usually achieve low-quality results 
on monitoring scenes. 
Furthermore, for UDA from SYNTHIA to Cityscapes, HRDA \cite{hoyer2022hrda} 
achieves a 65.8\% mIoU score. 
However, HRDA only achieves a 45.4\% mIoU score, which also indicates 
the large domain gap between the driving scenes and the monitoring scenes.
Providing a high-quality human-labeled dataset for analyzing the effectiveness of different methods 
of monitoring scenes will be beneficial for the community. 
It enables the researchers to cross-validate the effectiveness 
of the segmentation methods, instance segmentation methods, and 
unsupervised segmentation methods on the traffic monitoring scenes.




\begin{table*}[t]
    \centering
    \small
        \caption{Evaluation results of previous scene parsing approaches
        on the TSP6K validation and test sets. 
        }\label{tab:seg_bench}
        \vspace{-8pt}
    \renewcommand{\arraystretch}{1.1}
    \setlength{\tabcolsep}{2.2mm}{
    \begin{tabular}{ r | r |c|c|c|c|c|c|c} \toprule[1pt] 
        \multirow{2}*{Methods}   & \multirow{2}*{Publication} & \multirow{2}*{Backbone}  & \multirow{2}*{Parameters}  
        & \multirow{2}*{GFlops}  & \multicolumn{2}{c}{Validation}   & \multicolumn{2}{c}{Test} \\ \cmidrule(lr){6-7} \cmidrule(lr){8-9}
         &  &  & &  & mIoU (\%) & iIoU (\%) & mIoU (\%)  & iIoU (\%) \\ \midrule[0.8pt]
        FCN \cite{long2015fully}                   &  CVPR'15   &  R50     & 49.5M  &  454.1    &   71.5 &  55.2   & 72.5 & 55.1 \\
        PSPNet \cite{zhao2016pyramid}              &  CVPR'16   &  R50     & 49.0M  &  409.8    &   71.7 &  54.8   & 72.6 & 54.8 \\
        DeepLabv3 \cite{chen2017rethinking}        &  ArXiv'17  &  R50     & 68.1M	&  619.3    &   72.4 &  55.0   & 73.3 & 55.0 \\
        UperNet \cite{xiao2018unified}             &  ECCV'18   &  R50     & 66.4M  &  541.0    &   72.4 &  55.2   & 73.1 & 55.0 \\
        DeepLabv3+ \cite{chen2018encoder}          &  ECCV'18   &  R50     & 43.6M  &  404.8    &   73.1 &  56.1   & 73.9 & 56.3 \\
        PSANet \cite{zhao2018psanet}               &  ECCV'18   &  R50     & 59.1M  &  459.2    &   71.3 &  54.5   & 72.6 & 54.8 \\
        EMANet \cite{li2019expectation}            &  ICCV'19   &  R50     & 42.1M  &  386.8    &   72.0 &  55.5   & 72.9 & 55.5 \\
        EncNet \cite{zhang2018context}             &  CVPR'18   &  R50     & 35.9M  &  323.3    &   71.4 &  54.8   & 72.7 & 55.0 \\
        DANet \cite{fu2019dual}                    &  CVPR'19   &  R50     & 49.9M  &  457.3    &   72.3 &  56.0   & 73.1 & 56.1 \\
        CCNet \cite{huang2019ccnet}                &  ICCV'19   &  R50     & 49.8M  &  460.2    &   72.0 &  55.3   & 73.1 & 55.3 \\
        KNet-UperNet \cite{zhang2021k}             & NeurIPS'21 &  R50     & 62.2M  &  417.4    &   72.6 &  56.8   & 73.7 & 56.5 \\
        OCRNet \cite{YuanCW20}                     &  ECCV'20   &  HR-w18  & 12.1M  &  215.3    &   73.2 &  55.3   & 73.7 & 55.1 \\
        SETR \cite{zheng2021rethinking}            &  CVPR'21   & ViT-Large& 310.7M &  478.3    &   70.5 &  44.9   & 70.7 & 45.0 \\
        SegFormer \cite{xie2021segformer}          & NeurIPS'21 &  MIT-B2  & 24.7M  &  72.0     &   72.9 &  54.6   & 73.8 & 54.9 \\
        SegFormer \cite{xie2021segformer}          & NeurIPS'21 &  MIT-B5  & 82.0M  & 120.8     &   74.5 &  56.7   & 74.8 & 56.7 \\
        Swin-UperNet \cite{liu2021swin}            & ICCV'21    & Swin-Base& 121.3M & 1184.6    &  74.9  &  57.4   & 75.6 & 57.2  \\
        SegNeXt \cite{guo2022segnext}              & NeurIPS'22 & MSCAN-Base &27.6M &  80.2     &  74.6  &  57.3   & 75.4 & 57.2  \\
        SegNeXt \cite{guo2022segnext}              & NeurIPS'22 & MSCAN-Large&48.9M & 258.6     &  74.8  &  57.7   & 75.6 & 57.6  \\ \midrule
        \textbf{DRD (Ours)}          &  --        & \textbf{MSCAN-Base}  &\textbf{46.1M} & \textbf{90.1}     &  \textbf{75.8}  &  \textbf{58.4}   & \textbf{75.9} & \textbf{58.0}  \\
        \bottomrule[1pt]
        \end{tabular} }
\end{table*}

\section{Evaluating Segmentation Methods} \label{sec:benchmark}
%

\subsection{Implementation Details}
%
We run all the scene parsing methods 
based on a popular codebase, mmsegmentation \cite{mmseg2020}.
All the models are trained on a node with 8 NVIDIA A100 GPUs.
%
For CNN-based methods, the input size is set to 769$\times$769.
For transformer-based methods, the input size is 
set to 1024$\times$1024.
We train all the methods for 160,000 iterations 
with a batch size of 16.
As SETR \cite{zheng2021rethinking} consumes large GPU memories, 
the input size is set to 769$\times$769, 
and the batch size is set to 8.
Furthermore, we utilize the default data augmentations 
in mmsegmentation \cite{mmseg2020}.
We utilize the mIoU \cite{long2015fully} metric to 
evaluate the performance of the scene parsing methods.
As mentioned in \cite{cordts2016cityscapes}, 
the mIoU metric is biased to the object instances 
with large sizes. 
However, the monitoring  traffic scene is full of 
small traffic participants.
To better evaluate the instances of the traffic participants, 
we utilize the iIoU metric over all classes containing instances, 
following \cite{cordts2016cityscapes}.

\subsection{Performance Analysis} \label{sec:pa}
The evaluating results of different methods can be 
found in \tabref{tab:seg_bench}. 
The scene parsing methods can be roughly divided into 
several groups. 
In the following, we mainly discuss the methods using encoder-decoder structure, 
the self-attention mechanism and the transformer structure.

%

\myPara{Encoder-decoder structure} 
The methods based on the encoder-decoder structure utilize the high-resolution 
low-level features to refine the details of segmentation maps.
UperNet \cite{xiao2018unified} and DeepLabv3+ \cite{chen2018encoder} 
apply the encoder-decoder structure to the segmentation network.
Compared with DeepLabv3 \cite{chen2017rethinking}, DeepLabv3+ \cite{chen2018encoder} 
utilizes the high-resolution features can further improve the 
segmentation results by more than 0.6\% mIoU score and 
1\% iIoU score on both two sets. 
We observe that the encoder-decoder structure is very useful 
for small object segmentation, where the iIoU score is improved 
by a large margin.

\myParaP{Self-attention mechanism} is widely used 
in scene parsing methods.
Among the evaluated methods, EncNet \cite{zhang2018context}, DANet \cite{fu2019dual}, 
EMANet \cite{li2019expectation}, and CCNet \cite{huang2019ccnet} 
all utilize different kinds of self-attention mechanisms.
Most of them obtain superior performance than FCN.
Benefiting from the ability to model the long-range pixel dependence,  
the methods based on the self-attention mechanism can 
refine the final segmentation results and 
improve the performance.
Among them, we observe that EncNet does not have performance gain compared to FCN.
We analyze that EncNet utilizes the channel-wise self-attention 
mechanism to build the global context, which cannot preserve the 
local details well, especially for traffic monitoring scenes 
that contain different sizes of traffic participants.

%
%
%

\myParaP{Transformer structure} has been successfully 
applied to the computer vision tasks \cite{dosovitskiy2020image,carion2020end}, 
which often achieves better recognition results 
than the convolutional neural network structures.
%
%
SETR \cite{zheng2021rethinking}, Segformer \cite{xie2021segformer}, 
Swin-UperNet \cite{liu2021swin}, and SegNeXt \cite{guo2022segnext} 
all utilize the transformer structure as the backbone for scene parsing.
Among them, SETR achieves much worse parsing results, while 
other transformer structures obtain superior results than the 
convolutional backbone.
%
%
Among them, UperNet \cite{xiao2018unified} using Swin \cite{liu2021swin} backbone 
performs much better than the ResNet-50 \cite{he2016deep} backbone 
in terms of both the mIoU and iIoU metrics.
Moreover, compared with Swin-UperNet, SegNeXt obtains a similar performance 
with only about 20\% parameters and 7\% GFlops.
%
%

In summary, the encoder-decoder structure, spatial self-attention mechanism, 
and transformer structure are very useful strategies for improving 
the traffic monitoring scene parsing.
In \secref{sec:drd}, according to the strategies, we design a more powerful decoder 
that can further improve SegNeXt, which performs better than the Hamburger decoder \cite{geng2021attention}.

\begin{table}[t]
    \centering
    \small
        \caption{Evaluation results of previous instance 
        segmentation approaches on TSP6K validation and test sets. 
        We run all the methods based on the ResNet-50 \cite{he2016deep} backbone.
        \vspace{-8pt}
        }\label{tab:ins_bench}
    \renewcommand{\arraystretch}{1.10}
    \setlength{\tabcolsep}{2.4mm}{
    \begin{tabular}{l|c|c|c|c} \toprule[1pt] 
        \multirow{2}*{Methods}  & \multicolumn{2}{c|}{Validation} & \multicolumn{2}{c}{Test} \\ \cline{2-5}
        & \multicolumn{1}{c|}{$\mbox{AP}_{box}$} & \multicolumn{1}{c|}{$\mbox{AP}_{seg}$} 
        & \multicolumn{1}{c|}{$\mbox{AP}_{box}$} & \multicolumn{1}{c}{$\mbox{AP}_{seg}$} 
        \\ \midrule[0.8pt]
        YOLACT \cite{bolya2019yolact}         & 19.9 & 13.7 & 20.7 & 14.6 \\
        Mask-RCNN \cite{he2017mask}           & 27.2 & 23.5 & 27.0 & 23.5 \\
        SOLO \cite{wang2020solo}              & --   & 29.6 & --   & 29.8 \\
        SOLOv2 \cite{wang2020solov2}          & --   & 28.6 & --   & 28.6 \\
        QueryInst \cite{fang2021instances}    & 37.7 & 31.5 & 37.2 & 31.3 \\
        Mask2Former \cite{cheng2022masked}    & 32.9 & 31.3 & 32.5 & 31.4 \\
        \bottomrule[1pt]
        \end{tabular} }
\end{table}

\section{Evaluating Instance Segmentation Methods}
We provide each traffic image in TSP6K with 
additional instance annotations, which can be used to 
evaluate the performance of the instance segmentation methods 
on segmenting and classifying traffic participants (\ie humans 
and vehicles) in the traffic monitoring images.
The categories of the traffic participants are as 
follows: person, rider, car, truck, bus, motorcycle, 
and bicycle.
We evaluate several classic instance segmentation methods 
including  YOLACT \cite{bolya2019yolact}, Mask RCNN \cite{he2017mask}, 
SOLO \cite{wang2020solo}, SOLOv2 \cite{wang2020solov2}, 
QueryInst \cite{fang2021instances} and Mask2Former \cite{cheng2022masked}.
All the above methods are conducted based on 
the publicly available codebase, mmdetection \cite{chen2019mmdetection}.
The average precision (AP) metric is reported in \tabref{tab:ins_bench}.

\myPara{Performance Analysis} 
Among the evaluated methods, QueryInst \cite{fang2021instances} achieves 
superior performance than other methods.
It also achieves the best 7.4\% $\mbox{AP}_{s}$ score.
The poor performance indicates the existing methods struggle 
in small instance segmentation.
Furthermore, we observe that Mask-RCNN based on ResNet-50 achieves 40.9\% 
box AP and 36.4\% mask AP on Cityscapes, 
which exceed the models trained on TSP6K by more than 10\% AP.
The performance discrepancy demonstrates instance segmentation 
on TSP6K is still an enormous challenge.
We hope the additional instance annotations aid the community to 
improve the performance of the instance segmentation methods 
for segmenting the traffic participants in the monitoring scenes.

\begin{table}[h]
    \small
    \centering
    \caption{Evaluation of the unsupervised domain adaption methods.  }
    \label{tab:uda_bench}
    \vspace{-8pt}
    \renewcommand{\arraystretch}{1.05}
    \setlength\tabcolsep{1.0mm}
    \begin{tabular}{l|cc|cc} \toprule[1.0pt]
      \multirow{2}*{Methods}   & \multicolumn{2}{c|}{SYNTHIA $\rightarrow$ TSP6K}  & \multicolumn{2}{c}{Cityscapes $\rightarrow$ TSP6K}  \\ \cline{2-5}
                               & mIoU ($\%$) & Imprv ($\%$) & mIoU ($\%$) & Imprv ($\%$)\\   
       \midrule[0.8pt]
       Baseline                             & 21.7 &   0                   & 26.1 & 0  \\  
       ADVENT \cite{vu2019advent}           & 22.3 & \highlight{${+0.6}$}  & 31.7 & \highlight{${+5.6}$}\\ 
       DA-SAC \cite{araslanov2021self}      & 33.0 & \highlight{${+11.3}$} & 33.9 & \highlight{${+7.8}$}\\
       SePiCo \cite{xie2023sepico}          & 33.8 & \highlight{${+12.1}$} & 35.9 & \highlight{${+9.8}$}\\
       DAFormer \cite{hoyer2022daformer}    & 33.4 &\highlight{${+11.7}$}  & 39.5 & \highlight{${+13.4}$}\\ 
       HRDA \cite{hoyer2022hrda}            & 45.4 &\highlight{${+23.7}$}  & 54.1 & \highlight{${+18.0}$} \\
    \bottomrule[1.0pt]
    \end{tabular}  
    \vspace{-5pt}
\end{table}

\section{Unsupervised Domain Adaption} 
UDA methods for scene parsing are widely studied 
in recent years.
However, most UDA methods focus on adapting 
the synthetic driving scenes to the real driving 
scenes.
Benefiting from the proposed TSP6K dataset, we 
can study the UDA methods for adapting the 
driving scenes to the traffic monitoring scenes.
Specifically, we conducted UDA experiments for 
adapting SYNTHIA \cite{ros2016synthia} and Cityscapes \cite{cordts2016cityscapes} 
datasets to the TSP6K dataset, respectively. 
We select several classic UDA methods and evaluate them, 
including ADVENT \cite{vu2019advent}, DA-SAC \cite{araslanov2021self}, 
SePiCo \cite{xie2023sepico}, DAFormer \cite{hoyer2022daformer}, and 
HRDA \cite{hoyer2022hrda}.
The experiment results are shown in \tabref{tab:uda_bench}.
Note we only count and average the results of the common 
classes in both the source and target domains.

\myParaP{Performance Analysis:} We build a baseline, which 
trains DeepLab-v2 \cite{chen2017deeplab} on the source domain, 
and directly inferences on the target domain.
Compared with the baseline, all evaluated UDA methods 
outperform it by a large margin.
We can observe that the recent transformer-based UDA methods 
achieve better performance than CNN-based UDA methods.
The best performance of UDA methods is still far inferior 
to the fully-supervised methods (54.1\% \emph{vs} 72.4\%). 
Furthermore, UDA from Cityscapes to TSP6K achieves much higher 
performance than UDA from SYNTHIA to TSP6K.
This fact demonstrates the existing driving datasets 
can facilitate the traffic monitoring scene understanding.
We hope that the proposed dataset can facilitate the development 
of UDA methods for the task of traffic monitoring scene parsing.

\begin{figure*}[t]
  \centering
  \setlength\tabcolsep{1pt}
  \begin{overpic}[width=0.855\textwidth]{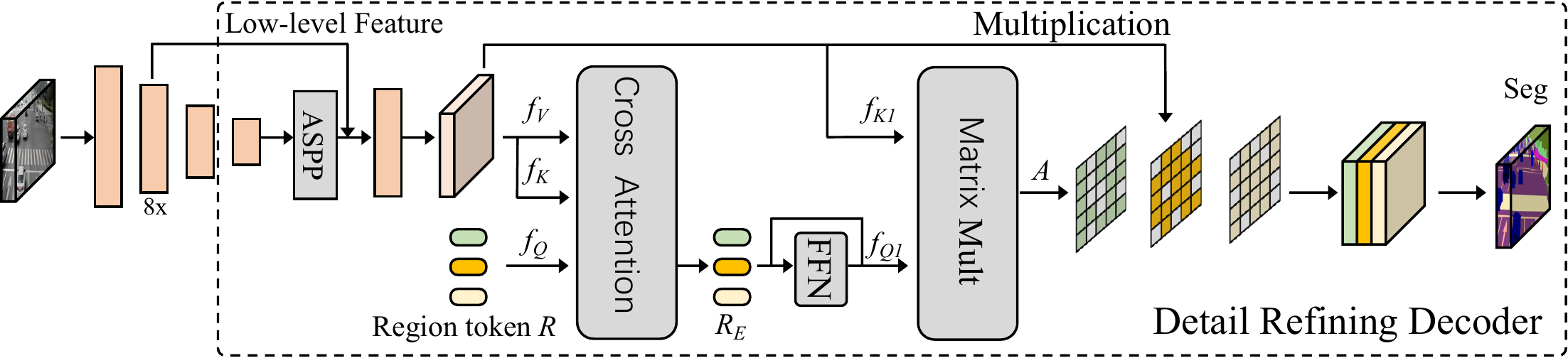}
  \end{overpic}
  \vspace{-2pt}
  \caption{Pipeline of the detail refining decoder. 
  Our decoder contains two parts. The first part is similar 
  to the decoder presented in DeeplabV3+~\cite{chen2018encoder}. 
  Differently, we use the feature maps from the third stage 
  ($\times8$ downsampling compared to the input) to fuse the feature maps from ASPP. 
  The second part is the proposed region refining module.
  }\label{fig:decoder}
  \vspace{-5pt}
\end{figure*} 

\section{Proposed Scene Parsing Method} \label{sec:drd}
As analyzed in \secref{sec:dataset}, the traffic monitoring scenes 
usually capture much more traffic content than the driving scenes 
and the scale and shape variances of different semantic regions are much larger. 
Moreover, small things and stuff take a large proportion.
These situations make accurately parsing the scenes challenging. 
To adapt to the traffic monitoring scenes, we propose a detail refining decoder.
The design principles of our decoder are two-fold.

\textbf{First}, as the spatial resolution of the last features 
from the backbone are very low, building decoders based on the low-resolution features 
usually generates coarse parsing results and, hence largely 
affects the small object parsing.
As verified in some previous works \cite{lin2017feature,wang2020solo,tian2020conditional}, 
the low-level high-resolution features are helpful for segmenting
small objects.
Thus, we utilize the encoder-decoder structure to fuse the low-resolution and
high-resolution features to improve the small object parsing.
\textbf{Second}, as analyzed in~\secref{sec:benchmark}, 
self-attention is an efficient way to encode spatial information
for scene parsing.
However, directly applying the self-attention mechanism 
to encode high-resolution features will consume massive computation 
resources, especially when processing high-resolution traffic scene images.
Inspired by \cite{cheng2022masked} that learns representations 
for each segment region, we propose to introduce several region tokens 
and build pairwise correlations between each region token and each
patch tokens from the high-resolution features.

\subsection{Overall Pipeline}

We construct the scene parsing network for traffic monitoring 
scenes based on the valuable tips summarized in~\secref{sec:benchmark}.
First, we adopt the powerful encoder presented in SegNeXt~\cite{guo2022segnext} 
as our encoder, 
which achieves good results with low computational costs on our TSP6K dataset.
Then, we build a detail refining decoder (DRD) upon the encoder.
The pipeline of the detail refining decoder 
is shown in \figref{fig:decoder}, which contains two parts.
For the first part, we follow the decoder design of DeepLabv3+~\cite{chen2018encoder} 
to generate fine-level feature maps.
The ASPP module is added to the encoder directly.
Note that we do not use the $\times$4 downsampling features 
from the second stage but the $\times$8 ones from the third stage 
as suggested in~\cite{guo2022segnext}.
The second part is the region refining module, which is described 
in the following subsection.

\subsection{Region Refining Module}
The region refining module is proposed to refine different semantic regions 
in the traffic image.
Formally, let $\mathbf{F}\in \mathbb{R}^{HW \times C}$ 
denote the flattened features from the first part of the decoder, 
where $H$, $W$, and $C$ denote the height, width, 
and the number of channels, respectively.
Let $\mathbf{R} \in \mathbb{R}^{N \times C}$ denote $N$ learnable region tokens, 
each of which is a $C$-dimensional vector. 
The flattened features $\mathbf{F}$ and the learnable region tokens $\mathbf{R}$ 
are separately sent into three linear layers to generate the query, key,
and value as follows:
\begin{equation}
\mathbf{R}_Q, \mathbf{F}_K, \mathbf{F}_V = f_Q(\mathbf{R}), f_K(\mathbf{F}), f_V(\mathbf{F}),
\end{equation}
where $f_Q(\mathbf{R})$, $f_K(\mathbf{F})$, and $f_V(\mathbf{F})$ are linear layers 
and $\mathbf{R}_Q \in \mathbb{R}^{N \times C}$, $\mathbf{F}_K \in \mathbb{R}^{HW \times C}$, 
$\mathbf{F}_V \in \mathbb{R}^{HW \times C}$.
We compute the multi-head cross-attention between $\mathbf{F}$ and $\mathbf{R}$ as follows:
\begin{equation}
  \mathbf{R}_E = \mathrm{Softmax} \left ( \frac{\mathbf{R}_Q \mathbf{F}_K^T}{\sqrt{C}} \right )  \mathbf{F}_V + \mathbf{R},
\end{equation} 
where $\mathbf{R}_E \in \mathbb{R}^{N \times C}$ is the resulting region embeddings. 
%
The region embeddings are then sent into a feed-forward network, 
which is formulated as:
\begin{equation}
    \mathbf{R}_O = \mathrm{FFN}(\mathbf{R}_E) + \mathbf{R}_E,
\end{equation}
where $\mathbf{R}_O$ is the output of the feed-forward network.
Here, following \cite{touvron2021going}, only the region tokens $\mathbf{R}_E$ 
are sent to the feed-forward block for an efficient process.

Next, $\mathbf{R}_O$ and $\mathbf{F}$ are delivered to two linear layers 
to generate a group of new queries and keys as follows:
\begin{equation}
  \mathbf{R}_{Q1}, \mathbf{F}_{K1} = f_{Q1}(\mathbf{R}_O), f_{K1}(\mathbf{F}).
\end{equation} 
We perform the matrix multiplication between $\mathbf{R}_{Q1}$ and $\mathbf{F}_{K1}$ to 
produce attention maps by 
\begin{equation}
    \mathbf{A} = \mathrm{Softmax} \left ( \frac{\mathbf{R}_{Q1} \mathbf{F}_{K1}^T}{\sqrt{C}} \right ),
\end{equation}
where $\mathbf{A} \in \mathbb{R}^{N \times HW}$ denotes $N$ attention maps and 
each attention map is associated with a semantic region.
When we attain the region attention maps, we combine $\mathbf{A}$ and $\mathbf{F} 
\in \mathbb{R}^{HW \times C}$ via broadcast multiplications, 
which can be written as follows:
\begin{equation}
    \mathbf{S}_{i,j,k} = \mathbf{A}_{i,j} \cdot \mathbf{F}_{j,k},
\end{equation}
where $\mathbf{S} \in \mathbb{R}^{N \times HW \times C}$ is the output.
Finally, $\mathbf{S}$ is permuted and reshaped to 
$\mathbb{R}^{N \times C \times H \times W }$, and then 
sent into a convolutional layer to generate the final segmentation maps.
%

\section{Experiments} \label{sec:scene_exp}
To verify the effectiveness of the proposed detail refining decoder, 
we conduct several ablation experiments.
Furthermore, we also compare our method with 
previous state-of-the-art methods.

\subsection{Ablation Study}
\myParaP{The number of region tokens and heads.}
First, we study the impact of the number of tokens and heads 
on the performance.
As shown in \tabref{tab:ablation2}, 
%
using 5 region tokens instead of 1 region token brings 
0.6\% mIoU scores and 1.2\% iIoU scores improvement.
This fact demonstrates that the number of region tokens 
largely affects the parsing of traffic participants, 
especially for small objects.
When further improving the number of region tokens, 
we observe nearly no performance gain, which indicates 
5 region tokens are enough for semantic region refining.
Besides, we also attempt to increase the number of attention heads.
It can be seen that adding more heads brings no 
performance gain.
For readers to better understand the region tokens, 
we have visualized the attention maps of different tokens, 
as shown in \figref{fig:attn}.
It can be seen that different tokens are responsible 
for different semantic regions. 

\begin{table}[h]
    \centering
    \small
        \caption{Ablation on the number of tokens and 
        attention heads.
        \vspace{-10pt}
        }\label{tab:ablation2}
    \renewcommand{\arraystretch}{1}
    \setlength{\tabcolsep}{1.80mm}{
    \begin{tabular}{c|cc|c|c} \toprule[1pt] 
        Settings &     \#Tokens              &       Attention Heads         &  $\mbox{mIoU}_{val}$ & $\mbox{iIoU}_{val}$ \\ \midrule[0.8pt]
         1  &        1                &        12            &   75.2  &  57.2 \\
         2  &        5                &        12            &   \textbf{75.8}   &  58.4\highlight{$_{(+1.2)}$} \\
         3  &        20               &        12            &   75.7 &  58.3\highlight{$_{(+1.1)}$} \\
         4  &        20               &        24            &   75.5  &  \textbf{58.6}\highlight{$_{(+1.4)}$} \\
        \bottomrule[1pt]
        \end{tabular} }
\end{table}

\newcommand{\addFig}[1]{\includegraphics[width=0.163\linewidth, height=0.10\linewidth]{figs/attention_maps/#1}}
\newcommand{\addFigs}[1]{\addFig{#1.jpg}  & \addFig{#1_0.png} & \addFig{#1_1.png} & \addFig{#1_2.png}  & \addFig{#1_3.png} & \addFig{#1_4.png} }
\begin{figure*}[t] 
    \centering
    \small
    \setlength\tabcolsep{0.6pt}
    \renewcommand{\arraystretch}{0.7}
    \begin{tabular}{ccccccc}
      \addFigs{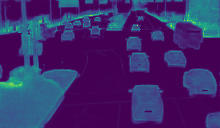} \\
      \addFigs{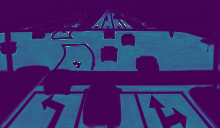} \\
      Image & $1^{th}$ token & $2^{th}$ token & $3^{th}$ token & $4^{th}$ token & $5^{th}$ token \\
    \end{tabular}
    \caption{Visualizations of the attention map corresponding to each token. 
    We randomly select several tokens for visualization. 
    One can see that the visualizations associated with different region tokens 
    focus on different semantic regions. 
    These region tokens can help our method better process the region details.}
    \label{fig:attn}
    \vspace{-2pt}
  \end{figure*}

\renewcommand{\addFig}[1]{\includegraphics[width=0.246\linewidth, height=0.14\linewidth]{figs/seg/#1}}
\renewcommand{\addFigs}[1]{\addFig{#1_gt1.jpg}   &  \addFig{#1_ccnet1.jpg}  & \addFig{#1_segnext1.jpg} & \addFig{#1_segnext_drd1.jpg} }
\begin{figure*}[t] 
    \centering
    \small
    \setlength\tabcolsep{0.6pt}
    \renewcommand{\arraystretch}{0.7}
    \begin{tabular}{cccc}
      \addFigs{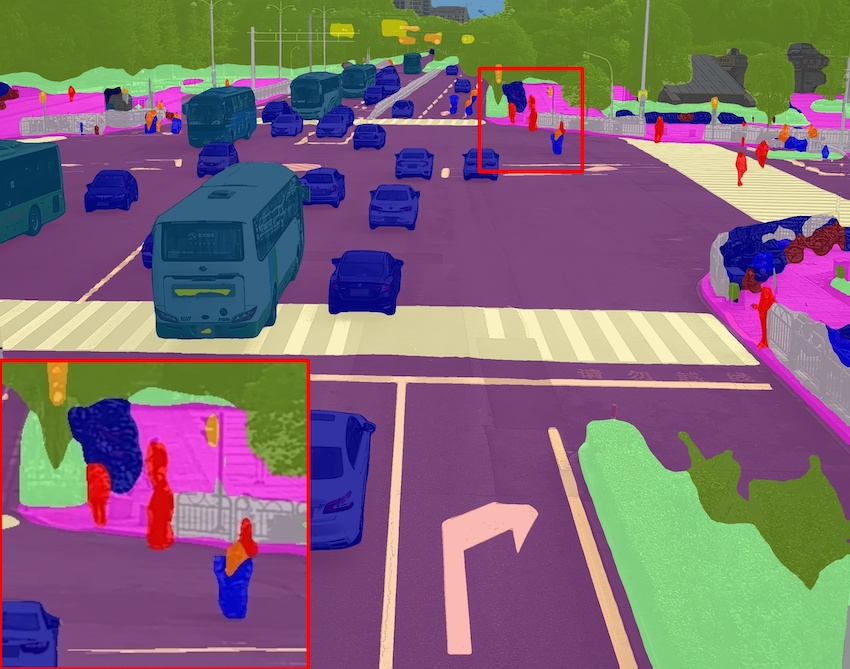} \\
      \addFigs{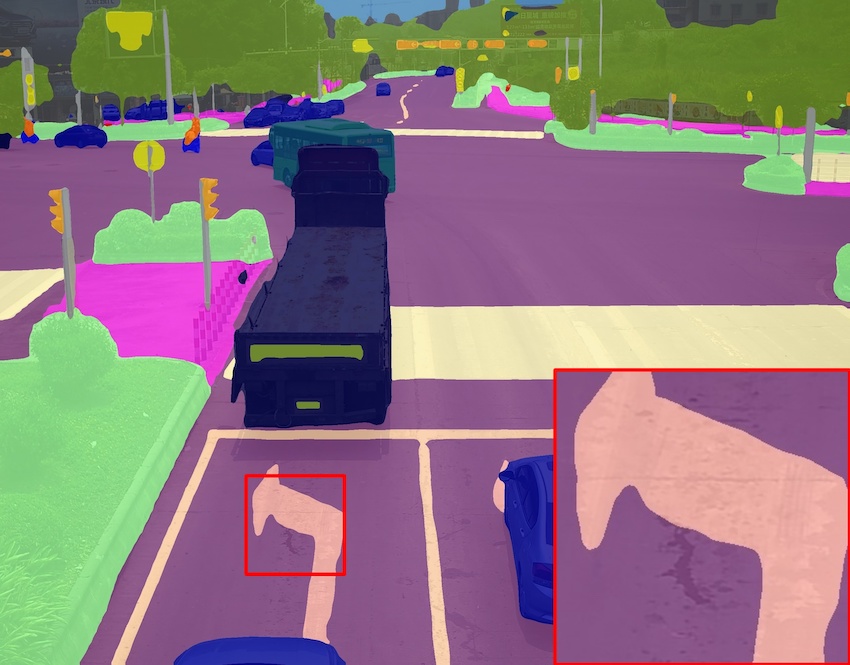} \\
      Ground-Truth  & CCNet \cite{huang2019ccnet} & SegNeXt \cite{guo2022segnext} & SegNeXt + DRD (ours) \\
    \end{tabular}
    \caption{Visualization of the scene parsing results from different methods. One can see that our method can well process the region details. When taking the bottom scene as an example, our method can generate a more accurate mask for the arrow while other methods fail.
    Zoom in for the best view. }
    \label{fig:seg}
    \vspace{-6pt}
  \end{figure*}

\myParaP{Region tokens \emph{vs.} Class Tokens.}
In the design of the detail refining decoder, we utilize the 
region tokens to refine a specific semantic region.
Here, one may raise a question: ``How would the performance go when 
we utilize class tokens as done in the original Transformers 
instead of the region tokens''?
We perform an experiment that learns 21 class 
tokens, each of which corresponds to a class.
The final concatenated features are sent to a depth-wise 
convolutional layer with 21 groups.
When using the class tokens, we can obtain 75.2\% mIoU scores 
and 57.1\% iIoU scores on the validation set.
Compared with class tokens, the decoder with 
5 region tokens can obtain 75.8\% mIoU and 58.4\% iIoU scores, 
which works better than using class tokens.
Moreover, when the number of classes in the dataset is large, 
the class tokens will consume high computational costs.
In contrast, using the region tokens is more flexible 
in that there is no need to adjust the number of region tokens 
when the number of classes rises.

\myParaP{The importance of the encoder-decoder structure.}
In Sec. 4.2 of the main paper, we have analyzed that the encoder-decoder 
structure is vital for small object parsing.
Thus, we apply the encoder-decoder structure to our segmentation network 
for utilizing high-resolution low-level features.
Without the encoder-decoder structure, i.e., we directly connect 
the region refining module to the encoder, 
the mIoU and iIoU scores decrease by 0.7\% and 1.3\%, respectively.
This experiment indicates that 
the high-resolution low-level features can benefit 
the parsing of the traffic participants.
Thus, the encoder-decoder structure is vital for scene parsing.

\subsection{Comparisons with Other Methods}
After performing a sanity check for the detail refining decoder, 
we compare the result of the proposed method with other methods 
on the TSP6K dataset.
\tabref{tab:seg_bench} lists the performance of different methods.
It can be seen that our method outperforms all 
previous methods and achieves the best results 
in terms of both two metrics.
The evaluation results demonstrate the effectiveness 
of the proposed decoder in parsing the traffic scenes.
Furthermore, we provide some qualitative results 
in \figref{fig:seg} for visual comparison.
We can see that our method can generate sharper results 
than CCNet \cite{huang2019ccnet} and SegNeXt \cite{guo2022segnext}.

\subsection{Traffic Flow Analysis}
\begin{figure}[t]
    \centering
    \setlength\tabcolsep{1pt}
    \begin{overpic}[width=0.45\textwidth]{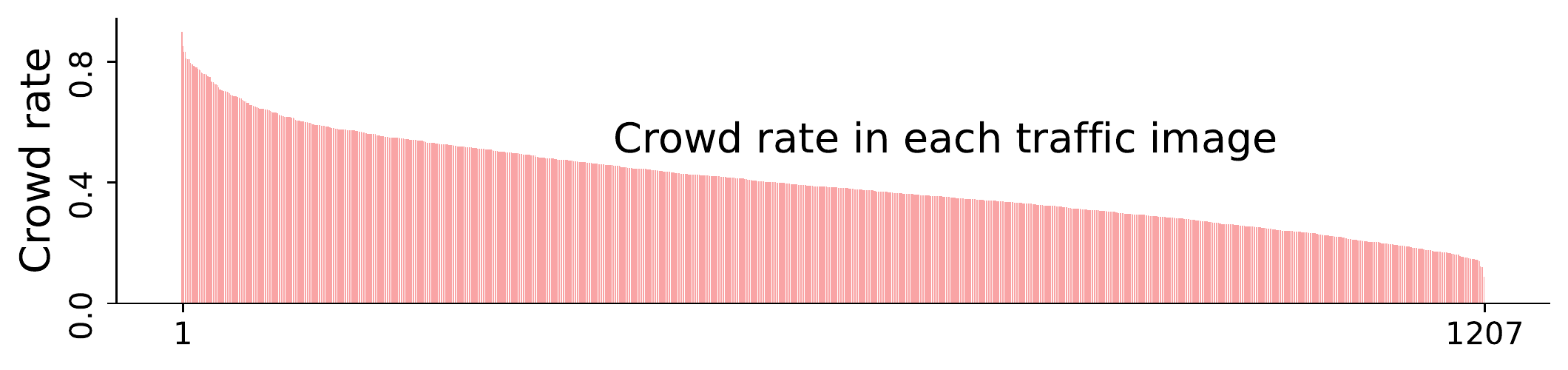}
    \end{overpic} \vspace{-10pt}
    \caption{Crowd rate analysis of TSP6K validation set.
    }\label{fig:crowded_rate}
    \vspace{-10pt}
\end{figure}

One underlying application scenario of the monitoring scene parsing models is analyzing the traffic flow.
Once we obtain the scene parsing results from 
well-trained models on the TSP6K dataset, we attempt 
to utilize these results to compute the traffic flow.
Our solution is very simple. 
We first compute the area of the traffic participants (humans and vehicles) $S_t$ 
and the area of the road $S_r$.
Then, the crowd rate can be approximately calculated by 
$S_t$ / ( $S_t$ +  $S_r$).
\figref{fig:crowded_rate} shows the crowd rate of 
different traffic images on the TSP6K validation set.
The higher the crowd rate, the more significant the traffic flow.
The forthcoming traffic participants can arrange 
their travel plans based on the current crowd rate.

\section{Conclusions}
In this paper, we have constructed the TSP6K dataset, 
focusing on the traffic monitoring scenes.
We have provided each traffic image with a semantic and instance label.
Based on the finely annotated TSP6K dataset, 
we have also evaluated a few popular scene parsing methods, 
instance segmentation methods, and UDA methods.
To improve the performance of the scene parsing, 
we design a detail refining decoder, which utilizes 
the high-resolution features from the encoder-decoder structure 
and refines different semantic regions based on the 
region refining module.
The detail refining decoder learns several region tokens 
and computes attention maps for different semantic regions.
The attention maps are used to refine the pixel affinity 
in different semantic regions.
Experiments have shown the effectiveness of 
the detail refining decoder.

\vspace{5pt}
\noindent \textbf{Limitations:}
The dataset contains 6,000 labeled images.
We still have a large amount of images remaining unlabeled, 
which can be further explored.
The dataset is not diverse geographically, which lacks scenes 
from the left-hand driving countries.
The modality of TSP6K only contains RGB images, 
which limits the development of multi-modal models.

{\small
\bibliographystyle{ieee_fullname}
\bibliography{tsp}
}

\end{document}